\documentclass[sn-basic]{sn-jnl}

\usepackage{graphicx}%
\usepackage{multirow}%
\usepackage{amsmath,amssymb,amsfonts}%
\usepackage{amsthm}%
\usepackage{mathrsfs}%
\usepackage[title]{appendix}%
\usepackage{xcolor}%
\usepackage{textcomp}%
\usepackage{manyfoot}%
\usepackage{booktabs}%
\usepackage{algorithm}%
\usepackage{algorithmicx}%
\usepackage{algpseudocode}%
\usepackage{listings}%
\usepackage{booktabs}
\usepackage{adjustbox}
\usepackage{cleveref}
\crefname{figure}{Fig.}{Figs.}

\theoremstyle{thmstyleone}%
\theoremstyle{thmstyletwo}%

\theoremstyle{thmstylethree}%

\begin{document}

\title[Article Title]{MonoMM: A Multi-scale Mamba-Enhanced Network for Real-time Monocular 3D Object Detection}


\author[1]{\fnm{Youjia} \sur{Fu}}\email{youjia\_fu@cqut.edu.cn}
\equalcont{These authors contributed equally to this work.}

\author[1]{\fnm{Zihao} \sur{Xu}}\email{Xu139@stu.cqut.edu.cn}
\equalcont{These authors contributed equally to this work.}

\author[1]{\fnm{Junsong} \sur{Fu}}\email{juns\_fu@stu.cqut.edu.cn}

\author[1]{\fnm{Huixia} \sur{Xue}}\email{xue\_xue@stu.cqut.edu.cn}

\author[1]{\fnm{Shuqiu} \sur{Tan}}\email{tsq@cqut.edu.cn}

\author*[2,3]{\fnm{Lei} \sur{Li}}\email{lilei@di.ku.dk}




\affil[1]{\orgname{Chongqing University of Technology}}
\affil*[2]{ \orgname{University of Washington}}
\affil*[3]{\orgname{University of Copenhagen}}


\abstract{Recent advancements in transformer-based monocular 3D object detection techniques have exhibited exceptional performance in inferring 3D attributes from single 2D images. By incorporating depth information and visual features from images, these methods enhance spatial awareness, thereby playing a crucial role in applications such as autonomous driving and augmented reality. However, most existing methods rely on resource-intensive transformer architectures, which often lead to significant drops in computational efficiency and performance when handling long sequence data. To address these challenges and advance monocular 3D object detection technology, we propose an innovative network architecture, \textbf{MonoMM}, a \textbf{M}ulti-scale \textbf{M}amba-Enhanced network for real-time \textbf{M}onocular 3D object detection. This well-designed architecture primarily includes the following two core modules: Focused Multi-Scale Fusion (FMF) Module , which focuses on effectively preserving and fusing image information from different scales with lower computational resource consumption. By precisely regulating the information flow, the FMF module enhances the model’s adaptability and robustness to scale variations while maintaining image details. Depth-Aware Feature Enhancement Mamba (DMB) Module : It utilizes the fused features from image characteristics as input and employs a novel adaptive strategy to globally integrate depth information and visual information. This depth fusion strategy not only improves the accuracy of depth estimation but also enhances the model’s performance under different viewing angles and environmental conditions. Moreover, the modular design of MonoMM provides high flexibility and scalability, facilitating adjustments and optimizations according to specific application needs. Extensive experiments conducted on the KITTI dataset show that our method outperforms previous monocular methods and achieves real-time detection.}




\maketitle

\section{Introduction}\label{sec1}

In 2D object detection technologies based on convolutional neural networks (CNNs) \cite{lin2017focal,ren2015faster,wang2023yolov7,wang2024yolov9} have developed rapidly. They are widely used in various fields, such as license plate recognition and defect detection. However, in areas such as robot navigation, autonomous driving, and multi-object tracking, obtaining detection results that are closer to the real world is difficult to achieve with 2D object detection. Therefore, to obtain detection results that are closer to reality, many researchers use distance sensors, e.g. LiDAR \cite{he2020structure,lang2019pointpillars,shi2020pv,shi2019pointrcnn,yang2024mixsup,li2023hierarchical,oehmcke2024deep,zhang2024research} or binocular cameras \cite{li2019stereo,peng2022side,shi2022stereo,wang2019pseudo} as input devices to acquire accurate depth information. Although high performance is achieved, these methods face high hardware costs. In contrast, obtaining high-precision 3D detection results from a single image can greatly reduce computation and equipment costs. Currently, some methods using only monocular cameras \cite{qian20223d,li2022diversity,brazil2019m3d,liu2020smoke} have been proposed, achieving promising results by utilizing the geometric constraints between 2D and 3D. However, since a single image cannot directly obtain depth information, the performance is far from satisfactory, thus attracting the attention of many researchers.

Depth estimation has witnessed substantial advancements, culminating in enhanced precision for monocular depth estimation models. Consequently, a plethora of researchers have leveraged sophisticated monocular depth estimation models to generate depth information, thereby augmenting monocular 3D object detection tasks. Pseudo-LiDAR-based methodologies \cite{ma2019accurate, weng2019monocular} transmute estimated depth maps into point clouds, simulating authentic LiDAR signals, and subsequently employ advanced LiDAR detectors for 3D object detection. Furthermore, fusion-based approaches \cite{ding2020learning, ouyang2020dynamic, wang2021depth} discretely extract and amalgamate image features and depth features for 3D object detection. While these methodologies exhibit proficiency in object localization, they are susceptible to the potential loss of salient features during the feature extraction process and are encumbered by inaccuracies inherent in the estimated depth images.

Pioneering research has expanded the application of Mamba to visual tasks, exemplified by VMamba \cite{zhu2024vision, wang2024mamba}, which enhances global effective receptive fields in both horizontal and vertical directions through selective scanning. Nonetheless, the utilization of Mamba in 3D monocular object detection remains relatively unexplored. This paper investigates the application of Mamba in monocular 3D object detection, optimizing its design to address the specific requirements of this task. Through targeted adjustments, we aim to establish a robust foundation for future research endeavors in this domain.

To address the issues identified, we propose a network for monocular 3D object detection named MonoMM. This network introduces a Focused Multi-Scale Fusion (FMF) module, which captures rich information across multiple scales using a specialized feature-focusing module with parallel depth convolutions. This approach facilitates the improved fusion of detailed contextual information from each scale, thereby enhancing 3D object detection performance. Furthermore, the FMF module is lightweight, reducing computation time compared to other multi-scale fusion methods \cite{quan2023centralized, lin2023scale, zhang2023attention}.

Unlike MonoDTR~\cite{huang2022monodtr}, this paper introduces the Depth-Aware Feature Enhancement Mamba (DMB) module, which is based on Mamba and models the image sequentially. This method linearly projects feature information into vectors, treating them as sequential data similar to text. By utilizing a bidirectional selective state space, the visual information is efficiently compressed, promoting complementarity between deep and shallow features, and enabling comprehensive image content capture. Moreover, this approach incorporates positional embeddings into the feature information, enhancing spatial perception and increasing the model's robustness in dense prediction scenarios.

Our contributions are summarized as follows:
\begin{itemize}
    \item This paper introduces a novel Focused Multi-Scale Fusion (FMF) module that employs feature focusing and diffusion mechanisms. These mechanisms distribute features with rich contextual information across different detection scales, effectively reducing noise and interference.
    \item We propose an innovative Depth-Aware Feature Enhancement Mamba (DMB) module, which is the first work to efficiently integrate image contextual information with Mamba. By using an adaptive strategy, it combines depth and visual information, thereby enhancing the performance of the detection head.
    \item Experimental results on the KITTI dataset demonstrate that our method outperforms most monocular approaches and achieves real-time detection. Furthermore, the FMF and DMB modules can be easily integrated into existing image-based frameworks, enhancing overall performance.
\end{itemize}

\section{Related Work}\label{sec2}
\subsection{Monocular 3D Target Detection}\label{work1}
Several existing monocular 3D object detection methods build on 2D object detectors \cite{ren2015faster,wang2024yolov9}. MonoRCNN \cite{wang2021depth} enhances 3D object detection by decomposing object distance into physical height and projected 2D height, modeling their joint probability distribution. MonoDLE \cite{ma2021delving} emphasizes the importance of accurate 2D bounding box estimation for predicting 3D attributes, identifying depth error as a critical limitation. M3D-RPN \cite{brazil2019m3d} introduces depth-aware convolutions to generate 3D object proposals constrained by 2D bounding boxes. MonoCon \cite{liu2022learning} incorporates auxiliary learning tasks to improve generalization performance. Monopair \cite{chen2020monopair} leverages spatial relationships between object pairs to enhance 3D positional information. MonoJSG \cite{lian2022monojsg} refines depth estimation using pixel-level geometric constraints. MonoFlex \cite{zhang2021objects} addresses long-tail object prediction issues with multiple depth estimators. PDR \cite{sheng2023pdr} simplifies this approach with a lighter architecture using a single perspective-based estimator. MonoGround \cite{qin2022monoground} incorporates a local ground plane prior and enriches depth supervision by sampling points around the object's bottom plane. MonoDDE \cite{li2022diversity} extends depth prediction branches based on keypoint information, emphasizing depth diversity. Despite these advances, purely monocular methods face challenges in precise object localization due to the lack of depth cues and the accumulation of errors in geometric constraints.

\subsection{Depth-assisted Monocular 3D Object Detection}\label{work2}
To achieve higher performance, many approaches leverage depth information to assist in 3D object detection \cite{huang2022monodtr,zhang2023monodetr,he2023ssd,ding2020learning,wang2021depth,you2019pseudo,zhou2023monoatt,reading2021categorical,yu2024credit}. Some methods \cite{ding2020learning,wang2021depth} employ advanced depth estimators and camera parameters to map images into 3D space, transforming them into pseudo-LiDAR data representations. These representations are then used with LiDAR-based 3D object detectors to enhance detection. Pseudo-LiDAR++ \cite{you2019pseudo} optimizes stereo depth estimation, leveraging sparse LiDAR sensors to improve 3D object detection in autonomous driving. D4LCN \cite{ding2020learning} and DDMP-3D \cite{wang2021depth} develop fusion-based methods that combine images and estimated depth with specialized convolutional networks. CaDDN \cite{reading2021categorical} learns class-specific depth distributions per pixel to construct Bird's Eye View (BEV) representations and recover bounding boxes from BEV projections. MonoDTR \cite{huang2022monodtr} uses LiDAR point clouds as auxiliary supervision for its Transformer and employs learned depth features as input queries for the decoder. MonoDETR \cite{zhang2023monodetr} predicts foreground depth maps using object labels for depth guidance. To enhance inference efficiency, MonoATT \cite{zhou2023monoatt} introduces an adaptive tokens Transformer that assigns finer tokens to more critical regions in the image.

\subsection{State Space Models}\label{work3}
State space models (SSMs), originating from classical control theory \cite{kalman1960new}, have recently gained prominence in deep learning due to their capability to manage long-term dependencies and sequential data. Innovations such as the Hippo model \cite{gu2020hippo}, which utilized polynomial highest power operator initialization, have enhanced SSMs' ability to capture long-term dependencies. The LSSL model \cite{gu2021combining} demonstrated the effectiveness of SSMs in handling these dependencies, though it faced challenges related to computation and memory efficiency. To address these issues, Gu et al. introduced the Structured State Space Sequence Model (S4) \cite{gu2021efficiently}, which reduced computational overhead through a normalized parameterization strategy, thus making SSMs more practical. S4 advanced deep-state space modeling for long-term dependencies. The S5 model \cite{smith2022simplified} incorporated MIMO SSM and efficient parallel scanning, while H3 \cite{fu2022hungry} narrowed the performance gap between SSMs and Transformer models in NLP. Recently, Mamba \cite{gu2023mamba} has shown superior performance in NLP with its advanced design, including selective mechanisms and optimized hardware usage, surpassing some Transformer models. Mamba efficiently handles long sequences, maintaining linear computational costs with sequence length, unlike Transformers, which require exponentially increasing resources. This efficiency makes Mamba a more effective solution for large-scale sequence data.

\section{Methods}\label{sec3}

\subsection{Overview}\label{subsec1}
\begin{figure*}[htbp]
    \centering
    \includegraphics[width=1\textwidth]{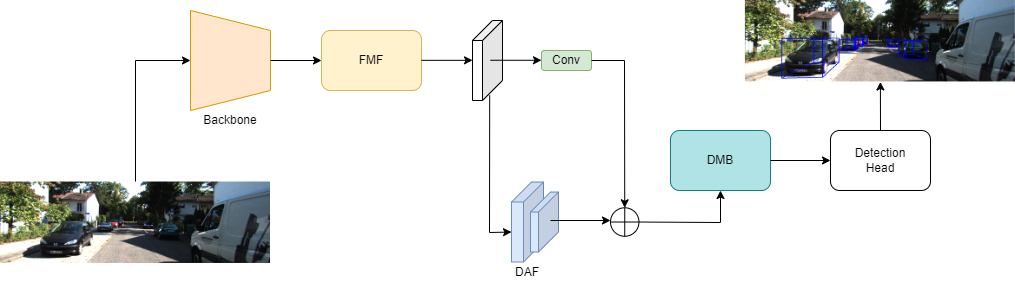}
    \caption{The general framework of MonoMM is proposed in this paper. First, the input image is processed by the backbone to extract features. The Focused Multi-Scale Fusion (FMF) module preserves detailed information at each scale through specific feature focusing and diffusion mechanisms. The Depth-Assisted Perception (DAP) module learns depth-aware features through assisted supervised learning. The Depth-Aware Feature Enhancement Mamba (DMB) module fully integrates visual information using adaptive strategies.}
    \label{fig:1}
\end{figure*}
\cref{fig:1} illustrates the MonoMM framework, which consists of five main components: a Backbone Network (Backbone), a Focused Multi-Scale Fusion (FMF) module, a Depth-Assisted Perception (DAP) module, a Depth-Aware Feature Enhancement Mamba (DMB) module, and a 2D-3D detection head. Following the approach in \cite{huang2022monodtr}, we adopt DLA-102 \cite{yu2018deep} as the Backbone Network. Given an RGB image input of size \(H_{in} \times W_{in}\), the Backbone Network outputs features at different layers. The FMF module integrates these feature maps to produce a fused feature map \(F \in \mathbb{R}^{C \times H \times W}\), where \(H = \frac{H_{in}}{8}\), \(W = \frac{W_{in}}{8}\), and \(C = 256\). Convolutional layers enhance feature representation by integrating spatial positional information across the image, thereby extracting higher-level and more abstract visual features. Additionally, we introduce the Depth-Assisted Perception (DAP) mechanism \cite{huang2022monodtr}, which employs multiple convolutional layers to specifically capture depth-related features. Subsequently, the DMB module efficiently integrates the visual and depth-aware features, further enhancing feature comprehensiveness and discriminability. To fully leverage these fused features, we utilize an anchor-based detection head architecture and select appropriate loss functions, aiming to achieve precise 2D and 3D object detection tasks simultaneously.

\subsection{Focused Multi-Scale Fusion Model}\label{subsec2}
\begin{figure*}[htb]
    \centering
    \includegraphics[width=1\textwidth]{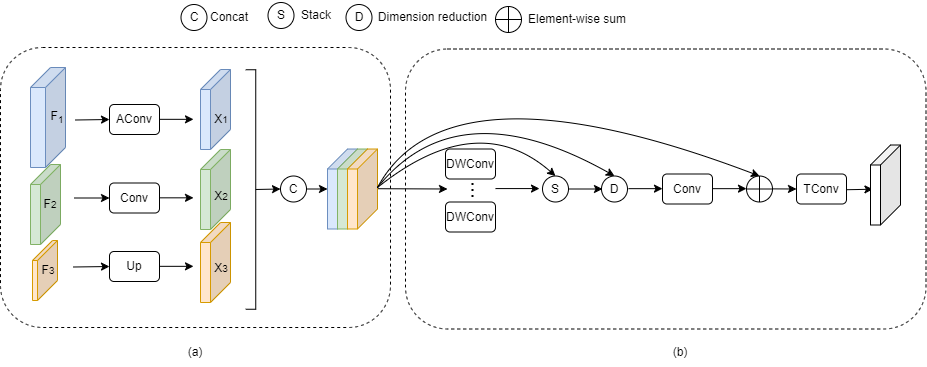}
    \caption{Feature Focusing Diffusion Model.The FMF module preserves more detailed information through two stages. (a) Multi-scale feature fusion generates feature F, (b) further refines the features and fuses them with F, finally generating F$_{out}$.}
    \label{fig:2}
\end{figure*}
Existing multi-scale feature fusion methods primarily focus on integrating contextual information across different layers. However, due to limitations in receptive field sizes, it is challenging to effectively fuse contextual information from different layers together. Moreover, incorporating attention mechanisms poses risks of increasing additional computational burden. To address these issues, inspired by the work of CFP \cite{quan2023centralized} and SMT \cite{lin2023scale} in dense prediction tasks, we propose a Global Focused Fusion (FMF) module for monocular image-based 3D object detection. In contrast to existing multi-scale fusion methods \cite{quan2023centralized,lin2023scale,yu2018deep}, our proposed FMF module not only captures long-range dependencies between different scales to comprehensively focus on integrating contextual information across layers but also avoids attention mechanisms, significantly reducing computational costs.

The initial fusion stage, depicted in ~\cref{fig:2}(a), consists primarily of three components: AConv \cite{wang2024yolov9}, a 1×1 convolutional layer, and an upsampling layer. Inside the AConv unit, two mechanisms are integrated: Average Convolution (AC) and Maximum Convolution (MC). The AC submodule effectively aggregates and smooths features by combining average pooling with convolution operations, reducing computational complexity and enabling efficient extraction of crucial features. Conversely, the MC submodule enhances feature capture of prominent features in the image by leveraging maximum pooling coupled with convolution, thereby improving the model's robustness to small-scale displacements. Specifically, different scale features extracted from the backbone are processed in parallel with the above operations to obtain rich multi-scale feature representations. This process enhances feature expressiveness, and by adjusting the size of feature maps, it effectively captures and retains rich information and details across various scales during the initial fusion stage. This comprehensive approach ensures a thorough grasp of both global and local features in the image. The resulting feature is denoted as $F \in R^{3C \times H \times W}$, where $C_1=256$,$H_1=\frac{H}{16}$,$W_1=\frac{W}{16}$. The process is illustrated in Equation~\ref{equ:1}.
    \begin{equation}
    \begin{aligned} 
        &X_i=Conv_i(F_i),i \in {1,2,3}, \\
        &F=Cat(X_i), i \in {1,2,3}
    \end{aligned}
    \label{equ:1}
    \end{equation}

In the detail fusion stage, the network focuses on deepening and refining the features generated in the initial fusion stage to produce more precise and detailed outputs. As shown in \cref{fig:2}(b), this stage begins by employing Depthwise Separable Convolution (DWConv) to parallelly process the initial fused features. This strategy delves deeper into feature extraction and effectively manages computational resource consumption, achieving deep-level feature fusion. Next, feature maps from various branch paths are stacked along the 0th dimension to ensure the complete preservation of multi-path information. To fuse this multi-path information into a unified feature map, the module adopts summation along the 0th dimension, significantly consolidating diverse information and subtle textures from different path transmissions, thereby constructing a rich and comprehensive feature representation for subsequent detection tasks. Subsequently, convolution and skip residual connection mechanisms are utilized to integrate the initial fused features, which have not undergone refinement, with the current features optimized through Depthwise Separable Convolution. This approach maintains the consistency and continuity of both global and local image features. Finally, the feature map dimensions are doubled by applying Transposed Convolution (TConv), effectively compensating for potential information loss introduced by previous downsampling steps. This step enhances resolution while ensuring detailed information's thorough restoration and transmission. The process is illustrated in Equation~\ref{equ:2}.

    \begin{equation}
        \begin{aligned} 
            F_{feature}&=Reduction(Stack(DW_i(F),F),F) \oplus F, \\
            F_{out}&=TConv(F_{[feature]}).
        \end{aligned}
        \label{equ:2}
    \end{equation}

\subsection{Depth-Aware Feature Enhancement Mamba Model}\label{subsec3}
\begin{figure}[htb]
    \centering
    \includegraphics[width=0.6\textwidth]{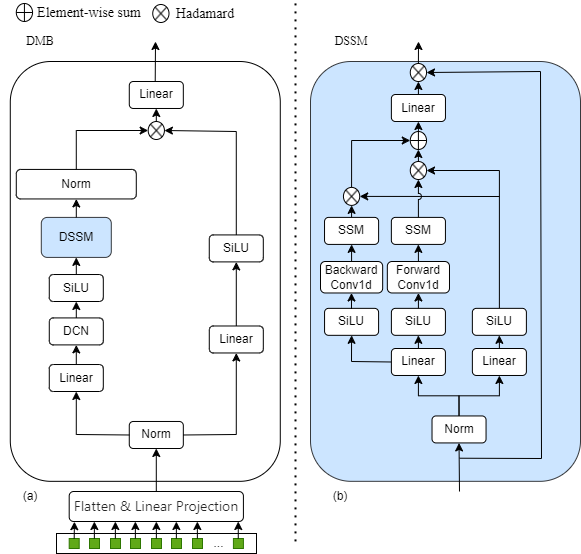}
    \caption{Depth-Aware Feature Enhancement Mamba Model.We first divide the input image into small patches and then project them onto patch embeddings. We then utilize Deformable Convolutional Networks (DCN) and DSSM to effectively learn the key features and contextual information within the image. This information is fused with another branch to effectively reduce the risk of gradient vanishing or explosion, while simultaneously enhancing feature learning capability.}
    \label{fig:3}
\end{figure}

To obtain depth-aware features for subsequent fusion with visual features, we drew inspiration from MonoDTR \cite{huang2022monodtr} and introduced the DAF module based on this approach. The DAF module effectively learns and refines depth-related feature information by incorporating precise depth maps as auxiliary supervision signals. This ensures that the DMB module comprehensively and deeply integrates depth-aware features with visual features, capturing and optimizing subtle information fused from both domains. As a result, it enhances the overall detection performance and accuracy.

The architecture of the DMB module we proposed is depicted in ~\cref{fig:3}(a). To handle the fused feature $F^d$ derived from the fusion of visual and depth-aware features, we first transform $F^d \in R^{C \times H \times W}$ into a flattened 2D sequence $x^p \in R^{p \times C}$, where $p={\frac{H}{{psize}^h}} \times \frac{W}{{psize}^w}$ denotes the total number of feature patches, and $C$ represents the number of channels. This transformation is formulated as shown in equation~\ref{equ:3}:
    \begin{equation}
    \begin{aligned} 
       T_0=[F_1^dW,...,F_p^dW]
    \end{aligned}
    \label{equ:3}
    \end{equation}
Where $F_p^d$ represents the p-th patch of depth-aware features, $W \in R^{p \times C}$ is a projection matrix with learnable characteristics. It maps these patches to a higher-dimensional feature space to capture more complex visual patterns. This approach draws inspiration from cutting-edge research such as ViT \cite{dosovitskiy2020image} and Vim \cite{zhu2024vision}, which innovatively apply patch-based methods by dividing images into continuous token sequences for global analysis. Following this concept, the patches segmented from $F^d$ are transformed into token sequences and fed into the DMB module, producing outputs $T_l$ after $l$ layers. To further refine and stabilize these feature representations, $T_l$ undergoes normalization to ensure consistency in the distribution of features across dimensions. This process is represented by Equation~\ref{equ:4}:
    \begin{equation}
    \begin{aligned} 
       T_l&=DMB(T_{l-1})+T_{l-1}\\
       f&=SiLU(Norm(_L^0))
    \end{aligned}
    \label{equ:4}
    \end{equation}

Specifically, after normalizing the input tokens, they are processed through two parallel branches. In the first branch, the sequence undergoes linear projection to transform it into a vector of dimension $E$, followed by a deformable convolution (DCN) to better capture and express the complexity of original information while reducing parameter count. Subsequently, it passes through a SiLU activation function, a DSSM layer, and normalization (Norm). In the second branch, the sequence also undergoes linear projection to transform it into a vector of dimension $E$, followed by a SiLU activation function. Afterward, features from both branches are aggregated through element-wise multiplication. $E$ represents the dimensionality of the hidden state. Finally, the features are projected back to the original size, generating output with the same shape as the input. The specific process is detailed in Algorithm.1:

\begin{algorithm}
\caption{DMB}\label{alg1}
\begin{algorithmic}[1]
\Require $F_p^d$,$W\in R^{p\times C}$
\Ensure $T^{\prime}$ 
\State $T$ = $W$ × $F_p^d$

\State $T_{norm}$ = Normalize($T$)

\State $D$ = LinearProjection($T_{norm}$,$E$)

\State $D$ = DCN($D$)

\State $D$ = SiLU($D$)

\State $D$ = DSSM($D$)

\State $D$ = Norm($D$)

\State $O$ = LinearProjection($T_{norm}$,$E$)

\State $O$ = SiLU($O$)

\State $F_{aggregated}$ = ElementWiseMultiplication($D$,$O$)

\State $T^{\prime}$ = LinearProjection($F_{aggregated}$,$original_{size}$)

\State Return $T^{\prime}$
\end{algorithmic}
\label{alg:DMB}
\end{algorithm}

In this context, the DSSM in this paper is inspired by the bidirectional information flow concept from Vim \cite{zhu2024vision}, which captures dependencies between local regions in a top-to-bottom order and captures global context and details in a bottom-to-top order, forming more comprehensive information transmission to improve performance further. As shown in ~\cref{fig:3}(b), in the forward branch of the first branch, features transform a linear layer to map them to a new feature space to enhance their expressive power. Subsequently, a 1D convolution layer with a SiLU activation function introduces non-linear characteristics, enhancing feature learning capability. This process is combined with the SSM layer to promote effective feature refinement jointly. The reverse branch executes the same processing sequence as the forward branch in the opposite direction, ensuring comprehensive communication and refinement of feature information, deepening the model's understanding of details. The second branch also begins with a linear mapping step to preserve more global coarse-grained information, which is crucial for avoiding potential gradient explosion during training. A SiLU activation function follows this to maintain the non-linear propagation of features. Features from both branches are then fused through Hadamard multiplication, an aggregation method that effectively combines feature strengths from different perspectives. Finally, the fused features are reshaped to their original dimensions, preparing them for processing in the next network layer. 

\subsection{Train loss and 2D-3D Detection}\label{subsec4}

We employed a single-stage detection method \cite{lin2023scale,zhang2022dimension}, which uses predefined 2D-3D anchor points to determine bounding boxes. Each anchor point is defined with both 2D bounding box parameters $[x_{2d},y_{2d},w_{2d},h_{2d}]$ and 3D bounding box parameters $[x_{p},y_{p},z,w_{3d},h_{3d},l_{3d},\theta]$. Here, $[x_{2d},y_{2d}]$ represents the center point of the 2D box, while $[x_p,y_p]$ denotes the projection of the 3D object center onto the image. $[w_{2d},h_{2d}]$ is the size of the 2D box, and $[x_{3d},y_{3d},l_{3d}]$ is the size of the 3D box. The parameter $z$ refers to the distance from the 3D object center to the camera, and $\theta$ represents the viewing angle. During training, all ground truth labels are mapped to the 2D space to compute Intersection over Union (IoU) with each 2D anchor. When IoU exceeds 0.5, the anchor is selected for optimizing and assigning to the corresponding 3D box.

Our output transformation method draws inspiration from Yolov5's\cite{yolov5} strategy, predicting a set of 2D parameters $[t_{x},t_{y},t_{w},t_{h}]$ and a set of 3D parameters $[t_{x},t_{y},t_{w},t_{h},t_{l},t_{z},t_{\theta}]$ for each anchor point. These parameters quantify the offsets for 2D and 3D bounding boxes and output a classification score cls for each category. By combining the anchor points' baseline positions with the network's predictions, we can accurately recover the bounding boxes of the targets.

Loss function. In this paper, the classification loss $L_{cls}$ and regression loss $L_{reg}$ are adopted using focal loss \cite{lin2017focal} and smooth L1 loss \cite{ren2015faster}, respectively. Specifically:

    \begin{equation}
    \begin{aligned} 
        L_{cls}(y,\widehat{y})&=-\alpha(1-\widehat{y}_x)^\gamma\log(\widehat{y}_x),\\
        L_{reg}(y,\widehat{y})&=
        \begin{cases}
        0.5(y-\widehat{y})            &{\lvert y-\widehat{y} \rvert < 0},\\
        \delta(\lvert y-\widehat{y} \rvert - 0.5\delta)                &\text{other.}
        \end{cases}
    \end{aligned}
    \label{equ:5}
    \end{equation}

Where $\alpha$ is the weight coefficient to balance positive and negative samples, adjusting the importance of different classes; $\gamma$ is the focusing parameter to reduce the weight of easily classified samples, allowing the model to focus more on hard-to-classify samples. $\delta$ is a threshold parameter: when the absolute error is less than $\delta$, the loss function switches to squared loss to ensure smooth gradient output and avoid significant disturbance to the model in case of small errors. When the absolute error is greater than or equal to $\delta$, absolute loss is used instead, which helps handle larger errors and mitigate the impact of outliers on model training.

For the depth loss $L_{dep}$, we follow the approach set by MonoDTR \cite{huang2022monodtr}, treating depth estimation as a classification task. We utilize the ground truth values of depth bins $\hat{D}$ generated from LiDAR. We employ the focal loss for this purpose, formulated as follows:
        \begin{equation}
    \begin{aligned} 
    L_{dep}=\frac{1}{\lvert P \rvert}\underset{p\in P}{\sum}{FL(D(p),\widehat{D}(p)}.
    \end{aligned}
    \label{equ:6}
    \end{equation}

wherein $P$ is a pixel region on the image with a valid depth label.

\section{Experiments}\label{sec4}
\subsection{Experimental setup}\label{1}

To validate the effectiveness of our proposed method, we selected the widely used KITTI 3D object detection dataset as our evaluation platform. This dataset comprises a total of 7,481 training images and 7,518 testing images. Adhering to the methodology outlined in \cite{chen20153d}, we partitioned the training set into two subsets: a training subset containing 3,712 images and a validation subset consisting of 3,769 images. Within this partitioning framework, we conducted ablation experiments to systematically analyze the contribution of each component to the enhancement of model performance. This rigorous experimental design ensures the reliability of the results and validates the soundness of our proposed methodology.

Regarding evaluation methods, whether for 3D object detection or bird's-eye-view (BEV) detection tasks, Average Precision (AP) is the key performance assessment metric. This study employs the AP at 40 recall positions metric to mitigate potential biases. In benchmark testing, detection difficulty is categorized into "easy," "moderate," and "hard" based on object size, occlusion level, and truncation. The methods are primarily ranked based on their 3D AP values under the "moderate" difficulty setting (Mod.), consistent with the KITTI benchmark. Additionally, following official guidelines, separate Intersection over Union (IoU) thresholds of 0.7, 0.5, and 0.5 are set for detecting car, cyclist, and pedestrian categories, respectively.

The specific operations during implementation are as follows: Following the setup in \cite{huang2022monodtr}, the Adam optimizer is employed to train the network model. The entire training process spans 100 epochs, with each batch containing 12 samples. The initial learning rate is set to 0.0001 and dynamically adjusted using cosine annealing. At each pixel position of the feature map, 48 anchor boxes are designed, covering 3 different aspect ratios (0.5, 1.0, 1.5). Additionally, 12 anchor boxes of varying height sizes are generated based on an exponential function, indexed from 0 to 15. For the parameters of the 3D anchor boxes, the mean and variance of the 3D ground truth bounding boxes in the training dataset are computed as reference statistics for the anchor boxes. To reduce inference time, the top 100 pixels of each image are cropped during preprocessing, and all images are uniformly resized to 288 × 1280 dimensions. During training, random horizontal flipping is applied for data augmentation. Predictions with confidence scores below 0.75 are discarded during inference, and a Non-Maximum Suppression (NMS) algorithm with an IoU threshold of 0.4 is used to eliminate redundant predictions.

\begin{table*}[!hbp]
\centering
\caption{Comparison of automotive category detection performance on the KITTI test set with current state-of-the-art methods. The best and second best are highlighted in bold and underlined, respectively.} 
\resizebox{\textwidth}{!}{
\begin{tabular}{c|c|ccc|ccc}
\hline
\multirow{2}{*}{\begin{tabular}{@{}c@{}}Method\end{tabular}} & \multirow{2}{*}{\begin{tabular}{@{}c@{}}Times(ms)\end{tabular}} & \multicolumn{3}{|c|}{\begin{tabular}{@{}c@{}}AP$_{3D}$@IOU=0.7\end{tabular}} & \multicolumn{3}{|c}{\begin{tabular}{@{}c@{}}AP$_{BEV}$@IOU=0.7\end{tabular}} \\
\cline{3-8}
& & Easy. & Mod. & Hard & Easy. & Mod. & Hard \\
\hline
MonoPSR & 200 & 10.76 & 7.25 & 5.85 & 18.33 & 12.58 & 9.91 \\
M3D-RPN & 160 & 14.76 & 9.71 & 7.42 & 21.02 & 13.67 & 10.23 \\
MonoPair & 60 & 13.04 & 9.99 & 8.65 & 19.28 & 14.83 & 12.89 \\
MoVi-3D & 45 & 15.19 & 10.90 & 9.26 & 22.76 & 17.03 & 14.85 \\
M3DSSD & - & 17.51 & 11.46 & 8.98 & 24.15 & 15.93 & 12.11 \\
D4LCN & 200 & 16.65 & 11.72 & 9.51 & 22.51 & 16.02 & 12.55 \\
MonoDLE & 40 & 17.23 & 12.26 & 10.29 & 24.79 & 18.89 & 16.00 \\
CaDDN & 630 & 19.17 & 13.41 & 11.46 & 27.94 & 18.91 & \underline{17.19} \\
MonoFlex & 30 & 19.94 & 13.89 & 12.07 & 28.23 & 19.75 & 16.89 \\
GUPNet & - & 20.11 & 14.20 & 11.77 & - & - & - \\
MonoGround & 30 & 21.37 & 14.36 & 12.62 & \textbf{21.99} & 20.47 & \textbf{17.74} \\
MonoDTR & 37 & \textbf{21.99} & \underline{15.39} & \underline{12.73} & \underline{28.59} & 20.38 & \ 17.14 \\
\hline
MonoMM(Ours) & 40 & 21.13 & \textbf{15.67} & \textbf{12.97} & 27.56 & \textbf{20.70} & 16.57 \\
\hline
\end{tabular}
}
\label{tab:main1} 
\end{table*}

\subsection{Main Results}\label{2}
In the results for the car category on the KITTI test set, as presented in Table~\ref{tab:main1}, the method proposed in this paper (MonoMM) is evaluated against other recent state-of-the-art monocular 3D detection methods. The findings indicate that MonoMM achieves superior performance in AP$_{3D}$ for moderate objects, which is the most critical metric in KITTI evaluations. This is particularly noteworthy as moderate samples generally exhibit smaller sizes, and prior algorithms \cite{brazil2019m3d,liu2020smoke,ding2020learning} have a propensity to incorporate excessive coarse-grained information during feature fusion, leading to detection inaccuracies. Conversely, MonoMM alleviates the issue of excessive coarse-grained information integration through the implementation of the FMF module, thereby focusing more on detailed information. This refinement results in significant performance improvements in the detection of moderate objects.

\begin{table*}[!htbp]
\centering
\caption{Detection performance of Car category on the KITTI validation set. We utilize bold to highlight the best results.} 
\resizebox{\textwidth}{!}{
\begin{tabular}{c|ccc|ccc|ccc|ccc}
\hline
\multirow{2}{*}{\begin{tabular}{@{}c@{}}Method\end{tabular}} & \multicolumn{3}{|c|}{\begin{tabular}{@{}c@{}}AP$_{3D}$@IOU=0.7\end{tabular}} & \multicolumn{3}{|c|}{\begin{tabular}{@{}c@{}}AP$_{3D}$@IOU=0.7\end{tabular}} & \multicolumn{3}{|c}{\begin{tabular}{@{}c@{}}AP$_{BEV}$@IOU=0.7\end{tabular}} & \multicolumn{3}{|c}{\begin{tabular}{@{}c@{}}AP$_{BEV}$@IOU=0.7\end{tabular}}\\
\cline{2-13}
& Easy. & Mod. & Hard & Easy. & Mod. & Hard & Easy. & Mod. & Hard & Easy. & Mod. & Hard\\
\hline
M3D-RPN & 14.53 & 11.07 & 8.65 & 20.85 & 15.62 & 11.88 & 48.53 & 35.94 & 28.59 & 53.35 & 39.60 &31.76 \\
MonoPair & 16.28 & 12.30 & 10.42 & 24.12 & 18.17 & 15.76 & 55.38 & 42.39 & 37.99 & 61.06 & 47.63 & 41.92 \\
MonoDLE & 17.45 & 13.66 & 11.68 & 24.97 & 19.33 & 17.01 & 55.41 & 43.42 & 37.81 & 60.73 & 46.87 & 41.89 \\
CaDDN & 23.57 & 16.31 & 13.84 & - & - & - & - & - & - & - & - & - \\
GUPNet & 22.76 & 16.46 & 13.72 & 31.07 & 22.94 & 19.75 & 57.62 & 42.33 & 37.59 & 61.78 & 47.06 & 40.88 \\
MonoFlex & 23.64 & 17.51 & 14.83 & - & - & - & - & - & - & - & - & - \\
MonoDTR & \underline{24.52} & \underline{18.57} & \underline{15.51} & \underline{33.33} & \underline{25.35} & \underline{21.68} & \underline{64.03} & \underline{47.32} & \underline{42.20} & \underline{69.04} & \underline{52.47} & \underline{45.90} \\
\hline
MonoMM(Ours) & \textbf{25.74} & \textbf{20.78} & \textbf{16.72} & \textbf{35.25} & \textbf{26.78} & \textbf{23.87} & \textbf{65.92} & \textbf{49.97} & \textbf{43.68} & \textbf{69.89} & \textbf{52.53} & \textbf{45.97} \\
\hline
\end{tabular}
}
\label{tab:main2} 
\end{table*}

In this paper, experiments were conducted on the car category of the KITTI validation set under different IoU thresholds and task conditions, as shown in Table~\ref{tab:main2}. The proposed method outperformed several image-only methods. Specifically, compared to MonoDTR \cite{huang2022monodtr}, our approach demonstrated advantages across easy, moderate, and hard settings at an IoU threshold of 0.5. The model significantly improved 3D Average Precision (AP$_{3D}$) by 1.22, 2.21, and 1.00 percentage points, respectively. Additionally, Average Precision in Bird's-Eye-View (AP$_{BEV}$) increased by 1.92, 1.43, and 2.19 percentage points, respectively. These enhancements underscore the model's superior accuracy and validate its robust performance and broad applicability in complex environments.

\begin{table}[!htbp]
\centering
\caption{Comparison of the pedestrian (Ped.) and cyclist (Cyc.) categories on the KITTI test set with a 0.5 IoU threshold with the current best method. We use bold and underlines to highlight the best and second-best results.} 
\begin{tabular}{c|ccc|ccc}
\hline
\multirow{2}{*}{\begin{tabular}{@{}c@{}}Method\end{tabular}} & \multicolumn{3}{|c|}{\begin{tabular}{@{}c@{}}AP$_{3D}$(Ped.)\end{tabular}} & \multicolumn{3}{|c}{\begin{tabular}{@{}c@{}}AP$_{3D}$(Cyc.)\end{tabular}} \\
\cline{2-7}
& Easy. & Mod. & Hard & Easy. & Mod. & Hard \\
\hline
D4LCN & 4.55 & 3.42 & 2.83 & 2.45 & 1.67 & 1.36 \\
MonoDLE & 9.64 & 6.55 & 5.44 & 4.59 & 2.66 & 2.45 \\
MonoPair & 10.02 & 6.31 & 5.53 & 3.79 & 2.12 & 2.04 \\
MonoFlex & 9.43 & 6.31 & 5.26 & 4.17 & 2.35 & 2.04 \\
CaDDN & \underline{12.87} & \underline{8.14} & \underline{6.76} & \textbf{7.00} & \underline{3.41} & \underline{3.30} \\
\hline
MonoMM(Ours) & \textbf{14.86} & \textbf{9.95} & \textbf{8.34} & \underline{6.82} & \textbf{3.82} & \textbf{3.75} \\
\hline
\end{tabular}
\label{tab:main3} 
\end{table}

Table~\ref{tab:main3} presents the results for the pedestrian and cyclist categories on the KITTI test set, further illustrating the performance of our model in these categories. Detecting pedestrians and cyclists is more challenging than the car category, primarily due to their smaller size and non-rigid body structure, making accurate localization more difficult. Overall, the model significantly outperforms other state-of-the-art methods in the pedestrian category, achieving approximately a 15\% improvement. Regarding 3D detection for cyclists, we also achieved results comparable to CaDDN \cite{reading2021categorical} in the easy difficulty and surpassed other methods in the moderate and hard difficulties. The results shown in Table 3 validate our model's exceptional versatility and scalability. Therefore, the method proposed in this paper enables accurate detection of objects with diverse appearances.

\subsection{Runtime Analysis}\label{3}

On a single Nvidia 4090 GPU, the entire validation set was processed with a batch size of 1, yielding an average runtime of 40 frames per second. This performance highlights the efficiency of our proposed approach. In comparison to the current state-of-the-art methods, our MonoMM model demonstrates a substantial speed enhancement, being 15 times faster than the CaDDN method \cite{reading2021categorical}. Several key factors contribute to this significant improvement in speed.

Firstly, CaDDN generates bird's-eye-view representations for 3D detection by predicting depth maps, which necessitates a more complex architectural design to achieve precise depth prediction, thereby resulting in increased processing times. Secondly, fusion methods typically deploy two separate backbone networks to extract both image and depth features, a process that inherently consumes more computational resources and time. Moreover, the inference time required for the depth estimator is an additional factor, not accounted for in Table~\ref{tab:main1}.

Conversely, the model proposed in this paper employs a lightweight FMF module, which is specifically designed to enhance detailed feature representation without incurring significant computational overhead. Additionally, the DMB module effectively integrates depth-aware features with visual features, substantially improving the model's feature processing capabilities while markedly reducing runtime. This strategic integration and efficient module design underscore the superior performance of our model.

\subsection{Ablation Study}\label{4}

To validate the effectiveness of the proposed components in this paper's model, We conducted ablation experiments as shown in Table~\ref{tab:main4}, where the model of [34] is taken as the baseline model. (a) Baseline.(b) Based on the baseline model, this paper replaces the feature fusion module with FMF to observe the impact of this feature fusion approach on the 3D target detection performance. (c)Based on the baseline model, the image features are sensed using the DMB module, and its impact on the results is observed using only image-sensing features. (d) Integration of contextual features using convolutional operations, summing them with depth-aware feature elements, and observing the results again. (e) MonoMM integrates all the above-mentioned modules.

\begin{table}[!htbp]
\centering
\caption{Ablation experiments, conducted on the KITTI validation set to study the effect of different modules on the car category.} 
\begin{tabular}{c|c|ccc}
\hline
\multirow{2}{*}{\begin{tabular}{@{}c@{}}Method\end{tabular}} & \multirow{2}{*}{\begin{tabular}{@{}c@{}}Ablation\end{tabular}} & \multicolumn{3}{|c}{\begin{tabular}{@{}c@{}}AP$_{3D}$@IOU=0.7\end{tabular}} \\
\cline{3-5}
& & Easy. & Mod. & Hard \\
\hline
a & Baseline & 19.35 & 15.47 & 12.83 \\
b & Baseline+FMF & 20.71 & 16.95 & 13.26\\
c & Baseline+DMB & 24.32 & 18.89 & 14.12 \\
d & Baseline+DMB+Conv & 25.03 & 20.97 & 14.94 \\
e & MonoMM & \textbf{25.74} & \textbf{20.78} & \textbf{16.72} \\
\hline
\end{tabular}
\label{tab:main4} 
\end{table}
As illustrated in Table~\ref{tab:main4}, the experimental results delineate the contributions and effects of each module on overall performance. According to the experiments, introducing the FMF  module into the baseline significantly improves the Average Precision of 3D detection (AP$_{3D}$) across different difficulty levels from 19.35/15.47/12.83 to 20.71/16.95/13.26, underscoring the positive impact of the FMF module in enhancing feature details and promoting detection performance. 

Integrating the DMB module into the baseline further elevates AP$_{3D}$ to 24.32/18.89/14.12. This substantial improvement indicates that the DMB module is markedly more efficient in integrating depth and visual information compared to the original configuration, thereby significantly boosting 3D object detection accuracy.

Moreover, combining the DMB module with convolutional layers in the baseline configuration results in an increase in AP$_{3D}$ to 25.03/19.62/14.14. This outcome highlights the effectiveness of convolutional operations in integrating contextual features and validates the DMB module as a robust strategy for the integration of visual and depth features.

Ultimately, the MonoMM model, which incorporates all proposed optimization modules, achieves substantial gains of 6.39/5.31/3.89 in AP$_{3D}$ over the baseline, thereby confirming the significant improvements in detection accuracy, model stability, and robustness as posited in this study. The aforementioned ablation experiments collectively demonstrate the rationality of the module designs and the efficacy of the proposed approach in addressing monocular 3D object detection tasks.

\begin{table}[!htbp]
\centering
\caption{Extending our model to other models} 
\begin{tabular}{c|ccc|ccc}
\hline
\multirow{2}{*}{\begin{tabular}{@{}c@{}}Method\end{tabular}} & \multicolumn{3}{|c|}{\begin{tabular}{@{}c@{}}AP$_{3D}$@IOU=0.7\end{tabular}} & \multicolumn{3}{|c}{\begin{tabular}{@{}c@{}}AP$_{BEV}$@IOU=0.7\end{tabular}} \\
\cline{2-7}
& Easy. & Mod. & Hard & Easy. & Mod. & Hard \\
\hline
M3D-RPN & 14.53 & 11.07 & 8.65 & 20.85 & 15.62 & 11.88 \\
M3D-RPN+Ours & \textbf{17.76} & \textbf{13.98} & \textbf{12.24} & \textbf{24.32} & \textbf{18.71} & \textbf{14.96} \\
Improvement & +3.23 & +2.91 & +3.59 & +3.47 & +3.09 & +3.08 \\
\hline
MonoDLE & 17.45 & 13.66 & 11.68 & 24.97 & 19.33 & 17.01 \\
MonoDLE+Ours & \textbf{19.32} & \textbf{16.54} & \textbf{14.38} & \textbf{28.52} & \textbf{22.88} & \textbf{20.82} \\
Improvement & +1.87 & +2.88 & +2.70 & +3.55 & +3.55 & +3.81 \\
\hline
\end{tabular}
\label{tab:main5} 
\end{table}

\begin{figure}[H]
    \centering
    \includegraphics[width=0.8\textwidth]{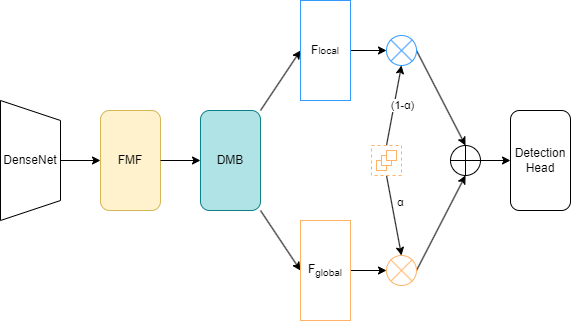}
    \caption{M3D-RPN after adding our module.M3D-RPN (Monocular 3D Region Proposal Network) is a framework designed for 3D object detection using a single camera image. }
    \label{fig:4}
\end{figure}

To validate the ease of integrating the modules into existing monocular 3D object detection models and to improve their performance, we selected two state-of-the-art models: M3D-RPN \cite{brazil2019m3d} and MonoDLE \cite{ma2021delving}. We integrated the proposed modules (FMF and DMB) into these models, resulting in the fused M3D-RPN model depicted in ~\cref{fig:4}. The practical steps involved the following operations: initializing with the features of these models (before the detection heads), processing these features using FMF and DMB to generate integrated features, and finally using the original model's detection heads to detect 3D objects using these integrated features. The core idea of this process is that our approach can be easily integrated ("plug-and-play") directly into existing models without requiring extensive modifications to their original structures. This flexibility is a significant advantage of our modules. As shown in Table~\ref{tab:main5}, with the assistance of FMF and DMB modules, these detectors achieved notable performance improvements on the KITTI validation set. This outcome demonstrates our approach's flexibility and underscores its applicability to various 3D object detection models. Furthermore, these enhancements illustrate that our method can improve performance without altering the original model architecture.

\subsection{Qualitative Results}\label{5}
\begin{figure*}[htb]
    \centering
    \includegraphics[width=1\textwidth]{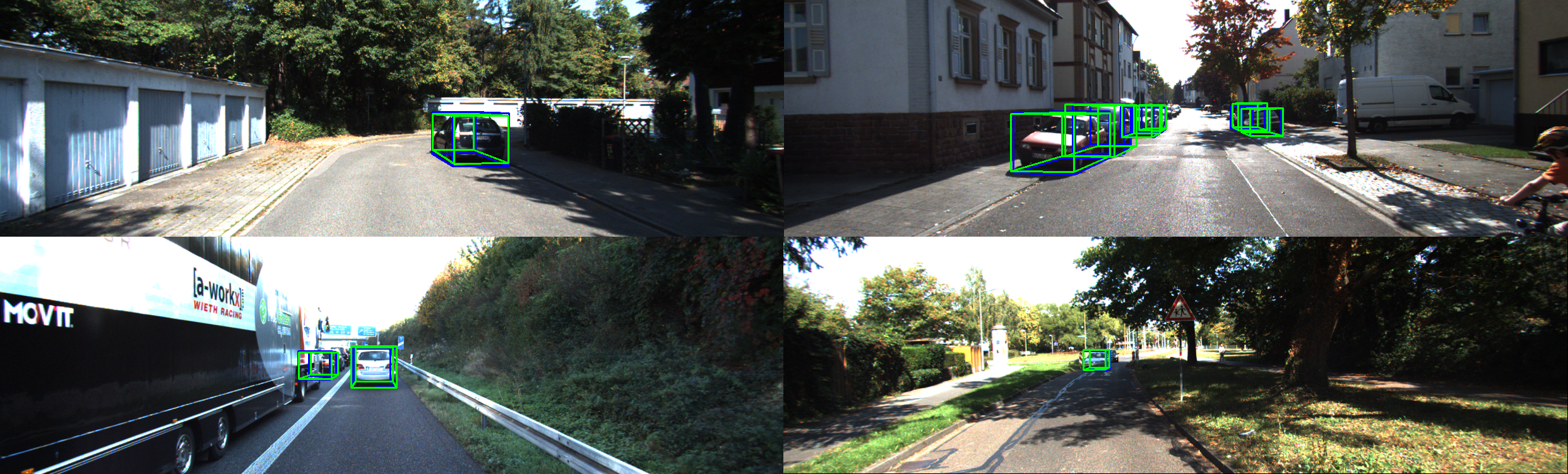}
    \caption{Qualitative examples on the KITTI validation set. We provide four images. In the images, the green boxes represent ground truth, and the blue boxes represent predictions from MonoMM.}
    \label{fig:5}
\end{figure*}
~\cref{fig:5} shows a qualitative example on the KITTI validation set that visually validates our approach. As can be seen from the figure, the predicted bounding boxes show a high degree of consistency and close fit between the predicted and actual labels, demonstrating a significant improvement in the accuracy of our model predictions. Finally, the model in this paper allows real-time inference. Our framework achieves a state-of-the-art trade-off between performance and latency.


\section{Conclusion}\label{sec5}

We introduce a novel algorithm for monocular 3D object detection, incorporating the Focused Multi-Scale Fusion Model (FMF). The FMF employs advanced feature focusing and diffusion mechanisms to disseminate context-rich features across multiple detection scales, thereby effectively mitigating noise interference. Additionally, this paper presents an innovative integration of the Depth-Aware Feature Enhancement Mamba (DMB) module into monocular 3D object detection, efficiently amalgamating contextual information from images. A pioneering adaptive strategy that combines depth and visual information significantly enhances the performance of subsequent depth prediction heads. This approach not only improves the accuracy of depth estimation but also optimizes model performance across various viewpoints and environmental conditions. Comprehensive experiments conducted on the KITTI dataset demonstrate that our model achieves real-time detection and surpasses the performance of existing monocular detection methods.

\backmatter
\bmhead{Acknowledgements}
the Chongqing Basic Research and Frontier Exploration Project (Chongqing Natural Science Foundation) under Grant CSTB2022NSCQ-MSX0918

the Science and Technology Research Project of Chongqing Education Commission (Youth) under Grant KJQN202101149

\section*{Declarations}
\bmhead{Competing interests}The authors declare that there is no conflict of interest in this study.

\bmhead{Ethics approval and consent to participate}There were no ethical issues in this study.

\bmhead{Data availability}Selected data supporting the results of this study are available from the authors upon request.


\bibliography{sn-bibliography}

\end{document}